\documentclass[a4paper,twoside]{article}

\usepackage{epsfig}
\usepackage{subfigure}
\usepackage{calc}
\usepackage{amssymb}
\usepackage{amstext}
\usepackage{amsmath}
\usepackage{amsthm}
\usepackage{multicol}
\usepackage{pslatex}
\usepackage{apalike}
\usepackage{SCITEPRESS}   

\begin{document}

\title{SentiCite \subtitle{An Approach for Publication Sentiment Analysis}}

\author{\authorname{Dominique Mercier\sup{1}, Akansha Bhardwaj\sup{2}, Andreas Dengel\sup{1,2}, Sheraz Ahmed\sup{2}}
\affiliation{\sup{1}Technical University of Kaiserslautern, Kaiserslautern, Germany \\\sup{2}German Research Center for Artificial Intelligence, Kaiserslautern, Germany}
\email{mercier@rhrk.uni-kl.de, firstname.lastname@dfki.de}
}


\abstract{With the rapid growth in the number of scientific publications, year after year, it is becoming increasingly difficult to identify quality authoritative work on a single topic. Though there is an availability of scientometric measures which promise to offer a solution to this problem, these measures are mostly quantitative and rely, for instance, only on the number of times an article is cited. With this approach, it becomes irrelevant if an article is cited 10 times in a positive, negative or neutral way. In this context, it is quite important to study the qualitative aspect of a citation to understand its significance. This paper presents a novel system for sentiment analysis of citations in scientific documents (SentiCite) and is also capable of detecting nature of citations by targeting the motivation behind a citation, e.g., reference to a dataset, reading reference. Furthermore, the paper also presents two datasets (SentiCiteDB and IntentCiteDB) containing about 2,600 citations with their ground truth for sentiment and nature of citation. SentiCite along with other state-of-the-art methods for sentiment analysis are evaluated on the presented datasets. Evaluation results reveal that SentiCite outperforms state-of-the-art methods for sentiment analysis in scientific publications by achieving a F1-measure of 0.71.}

\onecolumn \maketitle \normalsize \vfill


\section{\uppercase{Introduction}}
\label{sec:introduction}

\noindent Sentiment analysis is the process of computationally categorizing and identifying opinions present in a textual document or images. As a field, sentiment analysis has been gaining a lot of interest from the scientific community in recent years. Though some work has been done on various kinds of documents and text genres \cite{pang2004sentimental,whitelaw2005using,godbole2007large,pak2010twitter,agarwal2011sentiment,kouloumpis2011twitter,bahrainian2013sentiment,wu2015sentiment}, the focus of these approaches is on reviews, tweets, comments etc.

The main motivation for this work comes from the author's observation that there is an unavailability of a system capable of automatically analyzing the sentiment present in citations of scientific publications. Despite an immense need for such a qualitative approach, the existing scientometric approaches are quantitative and focus only on the number of times a paper is cited. The issue with such quantitative approaches is that they do not perfectly reflect the sentiment of a citation, i.e., if a paper is cited for its contribution or, as a bad example of research. In this context, sentiment analysis helps to identify relevant scientific publication. Though there are already few approaches existing in the field of sentiment analysis \cite{athar2011sentiment,xu2015citation,ma2016improve,yu2013automated}, none of them have defined an approach that focuses on the combination of sentiment in citations along with the nature of references. This is quite important as it can help researchers in determining the quality of a document for ranking in citation indexes by an inclusion of sentiment in a weighted approach based on nature of references as well. The presented work promises to fill the gap in current research approaches. The presented approach is adaptable to scientific documents from other venues like ACM, IEEE, Springer etc. as well. For other areas the approach needs a more generalized training set.

SentiCite is capable of analyzing a complete scientific publication, extracting locations of citations/references in the publication, and then associating sentiment to every citation in the paper based on the text where it is cited, provides information about the number of times a paper is referred positively, negatively or neutrally and identifies the nature of citation i.e., the motivation with which a paper was cited, e.g., a citation is made with an intention of dataset reference, further reading, information reference, or usage reference. Detecting nature and sentiment of citation helps in analyzing the publication from different dimensions simultaneously. 

Furthermore, this paper also presents two datasets (SentiCiteDB and IntentCiteDB) for both of the above-mentioned tasks i.e., Sentiment analysis of citation and nature of citation. The datasets contain about 2,600 citations along with their ground truth for sentiment and nature of citation. Also, we compare existing sentiment analysis approaches and analyze the nature of citations, i.e., finding the intent with which a paper is cited. 

The remaining paper is structured as follows. \textit{Section \ref{sec:related}} presents existing approaches available for sentiment analysis. \textit{Section \ref{sec:approach}} explains the proposed approach. \textit{Section \ref{sec:datasets}} presents the created datasets. \textit{Section \ref{sec:evaluation}} presents an evaluation of the proposed method including a comparison with other state-of-the-art sentiment analysis systems. \textit{Section \ref{sec:conclusion}} is the conclusion.


\section{\uppercase{Related Work}}
\label{sec:related}

\noindent This section provides an overview and a comparison of existing approaches available for sentiment analysis. For each approach, the work-flow and the used dataset is presented.

Pang and Lee \cite{pang2004sentimental} proposed sentiment analysis of movie reviews, which is based on minimum cuts in a graph to classify text and make use of the information that two sentences which are next to each other may share the same sentiment. Therefore, they created formulas for the individual score of the text piece and the associated score and made cuts for each class. In contrast to this paper they used movie reviews and only two classes.

Whitelaw et al. \cite{whitelaw2005using} address the issue that most of the sentiment analysis approaches rely on the \lq bag of words' or the \lq semantic orientation' and proposed the need for a semantic analysis of attitude expressions. This was done using a taxonomy of attitude types and appraisal groups which express an attitude. They used a lexicon based approach for the adjectives and performed a two class classification on a movie dataset. SentiCite uses a lexicon as well.

Godbole et al. \cite{godbole2007large} proposed a sentiment analysis approach for News and Blogs. In contrast to other automated approaches, they used synonym and antonym queries with respect to the same polarity to expand their lexicon. Therefore, a function that decreases the significance if the distance between two words is larger and a function that calculates a trust score if there are flips of positive and negative sentiment in a path was introduced. 

Pak and Paruoubek \cite{pak2010twitter} proposed a sentiment analysis approach for a twitter corpus. For the feature extraction, they used the presence of an n-gram as a binary feature. For the classification task, a multinomial Naive Bayes classifier was used. In common with the presented approach this was a three class classification task but with a very different domain.

Agarwal et al. \cite{agarwal2011sentiment} performed sentiment analysis on a twitter corpus where they used Part of Speech(POS) specific features and a tree kernel for their system. The proposed system used an emoticon dictionary and an acronym dictionary. Also, SentiCite uses POS features.

Kouloumpis et al. \cite{kouloumpis2011twitter} introduced an algorithm that uses linguistic features. They used different features like sentiment scores of a sentiment lexicon, part-of-speech features, unigram and bigram features and micro-blogging features to train a classifier. They used a twitter dataset and a emoticon dataset.

Bahrainian and Dengel \cite{bahrainian2013sentiment} proposed a hybrid polarity detection system as well as an unsupervised polarity detection system. The aspect detector module used a list of words as input to get the aspects of the target and deleted competitors. The polarity detection module had a preprocessing module, a feature generator and an SVM classifier. They performed a two class classification on twitter data. Also, the presented approach uses a SVM classifier.

Wu et al. \cite{wu2015sentiment} presented a sentiment approach that labels short text pieces with the help of probabilistic topics and similarities. The similarity was computed with a similarity function that summed up the probability for each word to appear on each topic. Their approach was done with headlines.

\begin{table}[b]
\caption{Comparison state-of-the-art.}
\label{table:compareTab}
\centering
\scriptsize
\begin{tabular}{lll} \\
\hline
\bfseries System & \bfseries Results & \bfseries Dataset \\
\hline
\cite{whitelaw2005using} & Acc: 90.2\% & movie\\ 
\cite{pang2004sentimental} & Acc: 86.4\% & movie\\ 
\cite{godbole2007large} & R: 0.69, P: 0.84 & WordNet \\
\cite{wu2015sentiment} & Acc: 41.1\% & news \\
\cite{bahrainian2013sentiment} & Acc: 89.78\%&  tweets \\
\cite{kouloumpis2011twitter} & F-score: 0.62 to 0.83 & tweets\\
\cite{pak2010twitter} & Acc: 0.62 to 0.8 & tweets \\
\cite{agarwal2011sentiment} & Acc: 57\% to 61\% & tweets \\ 
\hline
\end{tabular}
\end{table}

Table \ref{table:compareTab} shows the results of different approaches. The best performance on movie reviews was reached by Whitelaw et al. \cite{whitelaw2005using} and the best score for twitter data was reached Bahrainian and Dengel \cite{bahrainian2013sentiment}. The following paragraph contains work which is close to the work presented in this paper with respect to the data on which the work was done.

Athar and Teufel \cite{athar2012context} did work which is close to this paper. They used a support vector machine to classify the sentiment of different citations and further introduced a corpus for sentiment analysis on scientific papers. However, they mention that the distribution of the classes in not even.

Also, Abu-Jbara et al. \cite{abu2013purpose} did research on the question how to get a qualitative measurement instead of the count to classify the importance of a paper. They addressed that there is the need of a citation purpose to get a better understanding of the importance of a citation as well as a polarity score.

Di et al. \cite{di2013towards} provided results in the area of the citation purpose. Their system classified the different text pieces into categories which reflect their nature. However, their system had much more classes than the presented and only a few classes were dominant.

Xu et al. \cite{xu2013using} described a method to use extra textual features to classify citations which helps to find the relevant citations and discard less important. Also, this approach calculates a strength value based on the additional feature.

Ding et al. \cite{ding2014content} in their paper discuss the importance of the citation motivation and a value for different papers to get a better understanding of the impact a paper has on the community. They state that the deeper knowledge about the citations is useful to decide which papers are good because the number of publications increases very fast.

A different approach was presented by Wan et al. \cite{wan2014all} in which their system does not classify the sentiment but gets strengths values for the citation to understand the impact of the citation on the paper.

Furthermore, Mohammad et al. \cite{mohammad2016semeval} in their paper described their approach for stance detection which is close to the sentiment analysis topic. The stance detection tries to find out if the author of a text is in favor or against the target. Therefore, there is a relation between the sentiment and the stance but also a difference. The sentiment tries to find out the opinion of the author and the stance tries to find out the favorability.


\section{\uppercase{SentiCite: The Proposed Approach}}
\label{sec:approach}

\noindent This section provides details on the presented system, SentiCite, for analyzing scientific documents. SentiCite classifies the references/citations in a scientific publication into positive, neutral and negative classes and identifies the nature of references on a sentence level of the document. 

\begin{figure}[t]
\centering
\includegraphics[width=0.8\columnwidth]{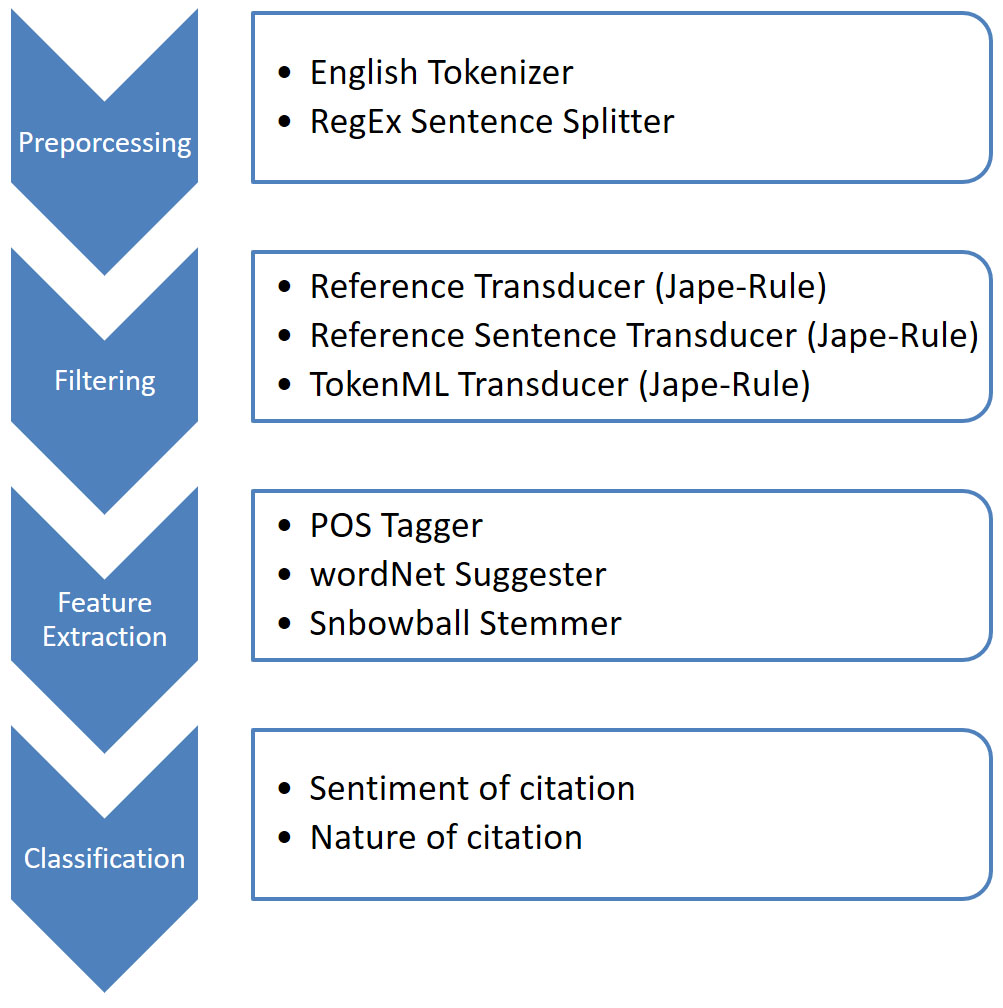}
\caption{Complete workflow for SentiCite.}
\label{fig_workflow}
\end{figure}

Figure \ref{fig_workflow} shows the four general steps and the corresponding sub-steps. SentiCite starts with a raw document and performs preprocessing to extract sentences. The extracted sentences are later filtered. Finally, a feature extraction and classification is performed. 

Furthermore, SentiCite also provides a visualization engine to show a multi-dimensional view of all citations in the paper with the detected sentiment of each citation and its nature.

\subsection{Preprocessing}

\noindent The first step of the SentiCite is to convert the scientific documents into a clean textual format by removing images, figures, and glyphs as these data has no influence on further processing steps. The next step is a \textit{\lq tokenization\lq} of the text followed by a sentence splitting method. This is done because the proposed method performs a sentence-level classification and these two steps are the basis for each further step.

\subsection{Filtering}

\noindent The purpose of this step is to filter the detected sentences. 
A sentence is filtered out if it does not have any citation/reference. It would be possible to use the neighbor sentences to validate the sentiment of the sentence with the citation but in this approach this is not done because it was assumed based on the manual labeling that the sentences with the citation contain the information that is mandatory for the sentiment assignment. This might be only the case because of the scientific area for which the approach was designed. The filtering is done in three sub-steps using regular expressions to identify the relevant information. These three steps identify the references, the sentences with references and the token within such sentences. 

\subsection{Feature Extraction}

\noindent The purpose of the feature extraction step is to get a meaningful set of features to train the classifiers. Besides the string of the tokens, three additional feature extraction modules were used. The POS tagger assigns different features e.g., type of the token, length or capitalization. The type of the token helps to indicate the importance of the token e.g., an adjective is more relevant than an article or nouns. Furthermore, additional features e.g. hypernyms or synonyms to cluster words are assigned with the help of wordNet\footnote{https://wordnet.princeton.edu/}. The last module is a stemming tool for a more generalized representation of the different tokens. 
Table \ref{table:differentFeatures} shows evaluation of different features. This procedure was done for the sentiment and the nature of citation. Furthermore, it has to be mentioned that features like n-gram, hyponyms, hypernyms and others were tested but the performance of these features was not sufficient. 

\subsection{Classification}
\noindent To finally classify sentences based on the extracted features, two different classifiers (i.e.,  Support Vector Machine (SVM) and a perceptron) are used, which leads to two different version of SentiCite i.e., SentiCite-SVM and SentiCite-Paum. It is important to mention that, for each classifier it was tested which subset of the features performs the best. An initial analysis showed that both versions are making different mistakes. Therefore, the results of both classifiers are fused to get SentiCite-Fusion method. This enables the combination of the results from both classifiers to achieve a more stable result. The performance increase obtained after using the SentiCite-Fusion approach is presented in the evaluation section in Table \ref{table:comparison}. This fusion approach checks both sentiment scores and decides based on a priority for the single classifiers which assignment is more likely to be correct this approach is called weighted fusion. The concrete technical background is that based on the performance of the classifiers during the evaluation of a subset a priority was created with the f-score of the classifiers. Therefore, the results of Table \ref{table:differentClasses} were used.

\subsection{Visualization}

\begin{figure}[b]
\centering
\includegraphics[width=0.9\columnwidth]{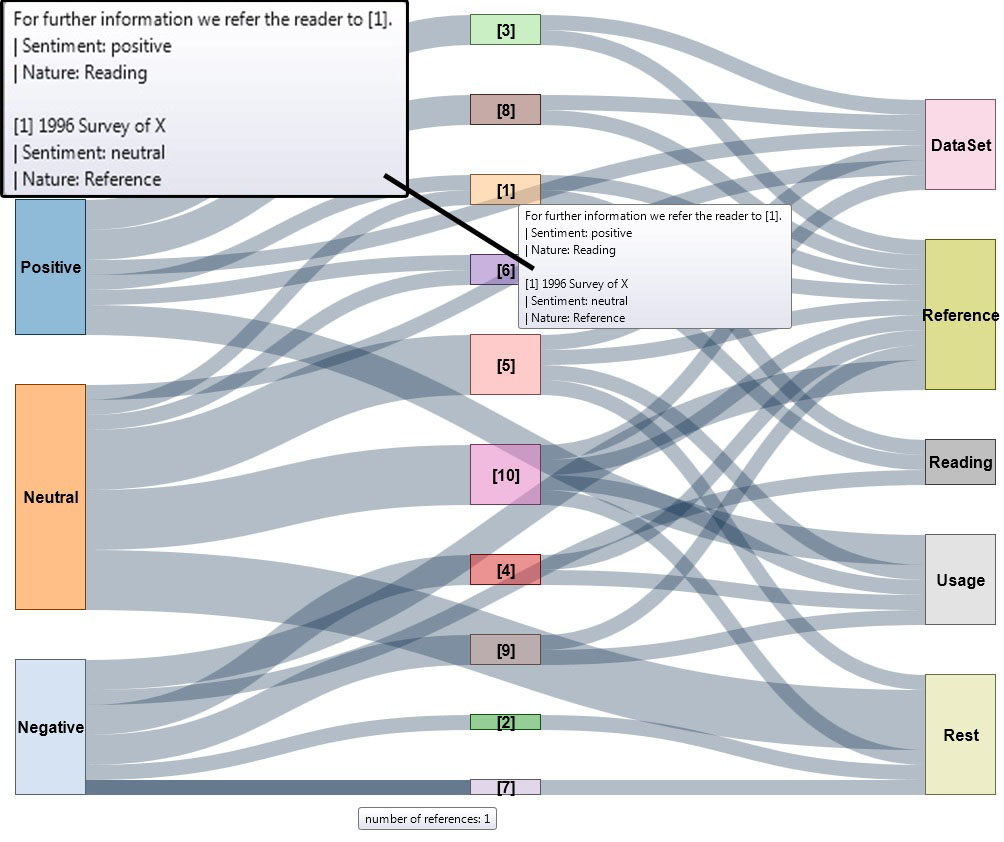}
\caption{Visualization of the system with sentiment and nature of references.}
\label{fig_visualisation}
\end{figure}

\noindent This paragraph provides an overview of SentiCite Visualization interface. Figure \ref{fig_visualisation} shows an analysis of a scientific publication with 10 references in the reference section. Each reference from the reference section is numbered in the middle. It is possible to get detailed information if the user hovers over the corresponding rectangle. The additional information includes the different sentences in which the given paper is referred and its corresponding sentiment and nature. In addition, the edges between the left column and the middle column show the number of positive, neutral and negative references if the user hovers over them. The width of the edges represents the count as well. Finally, the right column shows the nature of the references. Besides the visualization of the results as a chart, the user can also view extracted reference sentences, the references of the bibliography, the nature of each sentence and the sentiment of each sentence in the graphical interface of the system.


\section{\uppercase{Datasets}}
\label{sec:datasets}

\noindent This section provides details on the two new datasets introduced in this paper, i.e., SentiCiteDB and IntentCiteDB. These datasets were manually created for the evaluation of the SentiCite and have been compared with state-of-the-art algorithms. 

\subsection{SentiCiteDB}

\noindent SentiCiteDB is a dataset of publication sentiment analysis which is created using Scientific publications from International Conference on Document Analysis and Recognition (ICDAR) 2013. Sentences with citation are manually extracted from the publication and their manual ground truth is created. In total, SentiCiteDB contains about 2100 citations with 210 positive, 1805 neutral and 85 negative sentiment scores. Out of these, 50 citations from each of the three classes of sentiment are used for training of SentiCite. The statistics of the SentiCiteDB are shown in Table \ref{table:SentiCiteDB}. This dataset includes references from different sections of the document to make sure that the proposed method does not learn the sentiment classification for one specific section. The training set was selected with the same number of references for all classes to ensure that the classifier learns the classes in a sufficient way. Providing a real-world distribution a training set resulted in bad results for negative and positive labels. 

\begin{table}[b]
\caption{SentiCiteDB.}
\label{table:SentiCiteDB}
\centering
\begin{tabular}{llll}
\hline
& \bfseries Total & \bfseries Train set & \bfseries Test set \\
\hline
Positive & 210 & 50  & 160\\
Neutral & 1805 & 50  & 1755\\
Negative & 85 & 50  & 35\\
Overall & 2100 & 150  & 1983\\
\hline
\end{tabular}
\end{table}

\begin{table}[t]
\caption{Distribution of positive and negative references in different sections of publications.}
\label{table:heatmap}
\centering
\begin{tabular}{lll}
\hline
\bfseries Section & \bfseries Positive & \bfseries Negative \\ 
\hline
Introduction / Motivation & 0.22 & 0.28 \\ 
Information / Background & 0.2 & 0.06   \\ 
Related Work & 0.11 & 0.06 \\
Approach / Method & \textbf{0.3} & 0.1 \\
Evaluation / Experiments & 0.17 & \textbf{0.5} \\
\hline
\end{tabular}
\end{table}

A major challenge in the creation of this dataset was to find an equal number of sentences for each class because scientific documents include many more neutral references. For the testing of the classifier, 30 documents and their references were annotated manually. This reflects the real-world distribution of the different reference sentiments where papers have a huge number of neutral references and only a few are positive or negative. According to this real-world distribution  the classes are not equal distributed.

In addition to this dataset, different subsets were created with different numbers of references. These subsets include sets with an even distribution for the sentiment classes to evaluate the performance of the classifier for each class. Furthermore, a subset was created to compare the performance of the AYLIEN\footnote{https://developer.aylien.com/} sentiment system and the SentiCite. The developers state that the AYLIEN system is a state-of-the-art sentiment system.

To get a better understanding of the created datasets the structure of the test set was analyzed and a distribution of the classes is presented in Table \ref{table:heatmap}. This analysis presents a general behavior for this research area based on the dataset. This shows that the negative references are often in the evaluation section of a paper because the methods, in general, get outperformed by the proposed systems. 
In contrast to this, the positive references occur frequently in the proposed method section because authors adopt approaches which work well.

\subsection{IntentCiteDB}

\begin{table}[b]
\caption{IntentCiteDB.}
\label{table:IntentCiteDB}
\centering
\begin{tabular}{llll}
\hline
& \bfseries Total & \bfseries Train set & \bfseries Test set \\
\hline
Usage & 67 & 50  & 17\\
Reading & 57 & 50  & 7\\
Dataset & 65 & 50  & 15\\
Reference & 160 & 50  & 110\\
Rest & 162 & 50  & 112\\
Overall & 511 & 250  & 261\\
\hline
\end{tabular}
\end{table}

\noindent The same procedure was adopted for detecting the nature of references. IntentCiteDB was created the same way and in total contains 512 citations with five different classes:
The \textit{Usage} label describes a reference in which the cited authors algorithm is used. 
The \textit{Reading} label is used if the author refers to that paper for further information about something. 
The \textit{Dataset} label refers to references that are datasets which were used in the publication and listed.
The \textit{Reference} label is used for the bibliography.
The \textit{Rest} was used as a default label.    

It is possible to divide these classes into more meaningful subclasses but, in that case, a different training set would be required. As for the sentiment, each class has 50 training samples. The detailed statistics of this dataset and the class distribution according to the real-world distribution are shown in Table \ref{table:IntentCiteDB}. Like SentiCiteDB, the IntentCiteDB also contains references from all sections. For the testing, a subset of the used documents with 261 citations was used. This subset reflects the real-world distribution.


\section{\uppercase{Evaluation}}
\label{sec:evaluation}

\begin{table}[t]
\caption{Evaluation of different features for SentiCite (SC).}
\label{table:differentFeatures}
\centering
\begin{tabular}{lll}
\hline
\bfseries Label & \bfseries SC-SVM & \bfseries SC-Paum\\
\hline
Only POS & 0.7241 & 0.7336\\
Combination & \textbf{0.7260} & \textbf{0.8154}\\
\hline
\end{tabular}
\end{table}

\begin{table}[t]
\caption{Evaluation of different test corpus size for SentiCite (SC).}
\label{table:differentTestCorpusSize}
\centering
\small
\begin{tabular}{lllll}
\hline
\bfseries Approach & \bfseries 5 docs & \bfseries 10 docs  & \bfseries 20 docs  & \bfseries 30 docs\\
\hline
SC-SVM & 0.7111 & 0.7091 & 0.7141 & 0.7203\\
SC-Paum & 0.5727 & 0.7795 & 0.8218 & 0.8221\\
\hline
\end{tabular}
\end{table}

\begin{table}[t]
\caption{Evaluation of different training corpus size for SentiCite (SC). The number of references is the overall number and the class distribution is the real-world distribution.}
\label{table:differentTrainingCorpusSize}
\centering
\small
\begin{tabular}{lllll}
\hline
\bfseries Approach & \bfseries 25 refs & \bfseries 50 refs  & \bfseries 75 refs  & \bfseries 100 refs\\
\hline
SC-SVM & 0.6679 & 0.6775 & 0.6746 & 0.6603\\
SC-Paum & 0.7412 & 0.7659 & 0.7564 & 0.7621\\
\hline
\end{tabular}
\end{table}

\begin{table}[t]
\caption{Evaluation of different sentiment classes.}
\label{table:differentClasses}
\centering
\begin{tabular}{llll}
\hline
\bfseries Label & \bfseries SC-SVM & \bfseries SC-Paum & \bfseries SC-Fusion\\
\hline
Positive & \textbf{0.6173} & 0.3827 & 0.5556\\
Negative & \textbf{0.8333} & 0.7667 & \textbf{0.8333}\\
Neutral & 0.4020 & \textbf{0.6471}  & 0.4314\\
Overall & 0.5446 & \textbf{0.5634}  & 0.5352\\
\hline
\end{tabular}
\end{table}

\begin{table*}[t]
\caption{Ten run k-fold cross-fold for SentiCite (SC).}
\label{table:crossvalidation}
\centering
\begin{tabular}{llllllllllll}
\hline
\bfseries Algorithm & \bfseries F1 & \bfseries F2 & \bfseries F3 & \bfseries F4 & \bfseries F5 & \bfseries F6 & \bfseries F7 & \bfseries F8 & \bfseries F9 & \bfseries F10 & \bfseries Overall\\
\hline
SC-SVM & 0.47 & 0.6 & 0.53 & 0.67 & 0.6 & 0.47 & 0.47 & 0.6 & 0.4 & 0.53 & 0.53\\
SC-Paum & 0.4 & 0.53 & 0.47 & 0.6 & 0.6 & 0.67 & 0.53 & 0.53 & 0.4 & 0.6 & 0.53\\
\hline
\end{tabular}
\end{table*}

\noindent This section presents the results of the proposed approach and a comparison to state-of-the-art approaches. The experiments include results for the overall performance as well as for the different classes and the distribution of these classes.

Starting with the evaluation of the sentiment, the best F-score for the SentiCite-SVM was 0.7221 and 0.8327 for the SentiCite-Paum. To get these results the corpus of all 30 documents was used. 

It is important to note here, that, a classifier can often perform well on the neutral classes and bad on the positive and negative classes but, it can still result in a good overall accuracy because of the disproportionate ratio of neutral and positive or negative references in a scientific document. For more meaningful results further experiments were performed. Another interesting finding was that a combination of different POS, stemming and WordNet features performed very close to the POS features as shown in Table \ref{table:differentFeatures}. 

Furthermore, it has to be mentioned that other simple classifiers like a naive bayes, decision tree and a k-nearest-neighbor approach were tested too but not included in the evaluation section due to the bad performance compared to the support vector machine and the perceptron.

In the next experiment, the impact of the corpus size was tested. Therefore, the results are shown in Table \ref{table:differentTestCorpusSize}. The results show that for the SVM the f-score is almost the same for each run and that the corpus size has no impact. In addition, it shows that the perceptron performance increased in the beginning.

In addition to that, the impact of the training samples was calculated. Therefore, different numbers of references were used to train the classifiers. Table \ref{table:differentTrainingCorpusSize} shows that the results on a given test corpus are almost close for each of the training set sizes.

Finally, to validate that the results of the approach are correct a cross validation was done. The cross validation was done with the training corpus to have a balanced number of all classes. Therefore, the classifier was trained with a subset of the training corpus and evaluated on the remaining documents. This procedure as repeated 10 times with different splits. Also, this was the reason why the performance decreased because there was less training data. This was done because the training corpus was very large and the test corpus has for some labels only a few examples which was the case because of the distribution of the documents. The cross-validation was done exclusively with the training set which was split into two sets one for training and one for testing. This procedure was used ten times and the results in table \ref{table:crossvalidation} show that the performance is still good.

To get a better understanding of the results, Table \ref{table:differentClasses} shows the performances of the classifiers for the different classes on a collection of examples for each class. The results show the different performance of the algorithms for each class. For example, the SVM classifier performs well on the negative samples but the performance is worse on the neutral samples. The SentiCite-Paum performs quite well on the negative as well and slightly better on the neutral samples.

Table \ref{table:differentNature} shows the different nature of the analyzed references. As the numbers indicate, the classification of the different nature of references is not equally hard for the classifiers. The important labels have a good performance with scores around 0.667 for the datasets and up to 1.0 for the references.

\begin{table}[t]
\caption{Different nature of citation classes.}
\label{table:differentNature}
\centering
\begin{tabular}{llll}
\hline
\bfseries Label & \bfseries SC-SVM & \bfseries SC-Paum & \bfseries \#refs \\
\hline
Usage & 0.5294 & \textbf{0.8325} & 17  \\ 
Reference & \textbf{1.0} & 0.9455 & 110\\
Reading & 0.7143 & \textbf{0.8571} & 7 \\ 
Rest & 0.4533 & \textbf{0.4602} & 112  \\
Dataset & \textbf{0.667} & \textbf{0.667} & 15 \\
Overall & 0.7075 & \textbf{0.7099} & 261 \\
\hline
\end{tabular}
\end{table}

\begin{table}[t]
\caption{Comparison different classifiers and classes.}
\label{table:comparison}
\centering
\begin{tabular}{lllll}
\hline
\bfseries Method & \bfseries Pos & \bfseries Neg &  \bfseries Neu &  \bfseries F-Score\\
\hline
SentiCite-SVM & 11 & 9 & 18 & 0.65 \\
SentiCite-Paum & 8 & 9 & 20 & 0.63 \\
SentiCite-Fusion & 12 & \textbf{10} & 19 & \textbf{0.71} \\
AYLIEN & 0 & 2 & \textbf{23} & 0.36 \\
GoogleNLP & \textbf{13} & 0 & 16 & 0.4 \\
Correct values & 23 & 12 & 25 & \\
\hline
\end{tabular}
\end{table}

Finally, a comparison of the classifiers and state-of-the-art methods is shown in Table \ref{table:comparison}. The label shows the correct assignments of the classifiers for each label. For example, SentiCite-SVM assigned 11 positive references out of 23 in a correct way. It clearly shows that state-of-the-art systems like the AYLIEN\footnote{https://developer.aylien.com/} and the GoogleNLP only works for some classes and cannot deal with that scientific style of the citations. Furthermore, the results show that the state-of-the-art algorithm especially AYLIEN assigns neutral to almost every sample. An evaluation of the overall F-score shows that the proposed system's performances on this set were about 0.65 and 0.63. The AYLIEN software performance was 0.36. The same holds for the Google-NLP\footnote{https://cloud.google.com/natural-language/} API which was not able to classify the negative sentences and scored about 0.4. Most of the assignments produced by the Google-NLP were slightly positive or neutral. The best performance of this evaluation was 0.71 which was produced with the SentiCite-Fusion multi-classifier approach.

Table \ref{table:examplesSentiment} and Table \ref{table:examplesNature} show results of the sentiment and the nature of references assignment. Table \ref{table:examplesSentiment} shows short snippets of some reference sentences of the corpus, the human and the automatic annotation. The same holds for the Table \ref{table:examplesNature} which shows the results for the nature of references snippets.

This paragraph gives a summary of the findings presented in the evaluation section. The findings show that the use of part-of-speech features for a citation analysis is crucial and other features like a lexicon can increase the performance but that is not always the case. Therefore, a good selection of the features is important. The next finding was that the results for different classifiers differ for the different classes. Furthermore, the results show that a fusion approach can increase the overall performance and makes the system more robust. The third finding was the distribution of positive and negative sentiment in scientific papers which shows that positive citations are more likely to be in the method section and negative more likely to be in the evaluation section. Also, the classification of citations works with the nature of citations and finally the comparison with two state-of-the-art systems indicated that the proposed method which was trained on scientific papers outperformed the other systems on this specific task.

\begin{table*}[t]
\caption{Examples for sentiment assignment.}
\label{table:examplesSentiment}
\centering
\footnotesize
\begin{tabular}{lll}
\hline
\bfseries Reference & \bfseries Human & \bfseries System\\
\hline
However, [7] only focuses the disguise handwriting and this does not completely suffice .  & Neg & Neg \\ 
The most recent work in [12] achieves an accuracy of 61.9\% which is much lower  & Neg & Neg \\
These preprocessing steps have been described in greater detail in our prior work[7]. & Pos & Neu \\
The recent state-of-the-art of signature verification is summarized in [3]. & Pos & Pos \\
experiments with a context independent HMM-based system that uses a sliding window  [4]. & Neu & Neu \\
In [12], authors propose an SVM based active learning that utilizes the support vectors . & Neu & Pos \\
\hline
\end{tabular}
\end{table*}

\begin{table*}[t]
\caption{Examples for nature assignment.}
\label{table:examplesNature}
\centering
\scriptsize
\begin{tabular}{lll}
\hline
\bfseries Reference & \bfseries Human & \bfseries System\\ 
\hline
In the second configuration (SWT), the Stroke Width Transform [2] was used. & Usage & Usage \\
The first data set used in this thesis is the IAM database (IAMDB) 1 [10]. & Dataset & Dataset \\
For more details on the HMMs and the spotting system, we refer to [9]. & Reading & Reading \\
\ [1] E. Lleida et al., “Out-of-vocabulary word modeling and rejection for keyword spotting,” & Reference & Reference \\
On the other hand, forward-backward computing time is negligible with respect to that of CL generation [7]. & Rest & Rest \\ 
In this approach, as presented in [3], character HMMs are used to build both a “filler” model. & Usage & Usage \\ 
For the standard ICDAR 2011 dataset [14], the proposed method achieves state-of-the-art results in text localization. & Dataset & Rest\\
See [7], [3] for details about the meta-parameters of line-image preprocessing, feature extraction and HMMs. & Reading & Reading \\
\ [10] V. Romero et al., “Computer assisted transcript. & Reference & Reference \\
More recently, the same basic idea has also been used for KWS in handwriting images [3], [4]. & Rest & Reading \\
\hline
\end{tabular}
\end{table*}


\section{\uppercase{Conclusion}}
\label{sec:conclusion}

\noindent The evaluation results show that sentiment analysis of scientific documents depends on a good preparation and feature selection for the input. Some methods like the POS tagger, the removal of specific terms and the management of negations increase the performance of SVM classifiers. The SentiCite method shows that analyzing the sentiment of scientific papers with an SVM or a perceptron can produce good results. In addition, a multi-classifier approach based on SVM and perceptron can increase the performance. Furthermore, the SentiCite classifiers outperform all compared existing state-of-the-art methods. Classification of the nature of the references is also possible with the same approach. This shows that sentiment analysis can be done on an objective text as well and the results depend on the quality of the data and the processing. 

The information provided by the presented method has the potential to help the community to identify meaningful references and fill the gap in current citation indexes approaches. It is possible to classify scientific citations in a way that does not only depend on the number of citations and helps to get a better understanding of how important a citation is. A measurement that takes into account this information does not rely that strong on the number of citations and can rank new papers that have fewer citations higher than old ones.


\bibliographystyle{apalike}
{\small
\bibliography{mybib}}

\begin{thebibliography}{}

\bibitem[Abu-Jbara et~al., 2013]{abu2013purpose}
Abu-Jbara, A., Ezra, J., and Radev, D.~R. (2013).
\newblock Purpose and polarity of citation: Towards nlp-based bibliometrics.
\newblock In {\em Hlt-Naacl}, pages 596--606.

\bibitem[Agarwal et~al., 2011]{agarwal2011sentiment}
Agarwal, A., Xie, B., Vovsha, I., Rambow, O., and Passonneau, R. (2011).
\newblock Sentiment analysis of twitter data.
\newblock In {\em Proceedings of the workshop on languages in social media},
  pages 30--38. Association for Computational Linguistics.

\bibitem[Athar, 2011]{athar2011sentiment}
Athar, A. (2011).
\newblock Sentiment analysis of citations using sentence structure-based
  features.
\newblock In {\em Proceedings of the ACL 2011 student session}, pages 81--87.
  Association for Computational Linguistics.

\bibitem[Athar and Teufel, 2012]{athar2012context}
Athar, A. and Teufel, S. (2012).
\newblock Context-enhanced citation sentiment detection.
\newblock In {\em Proceedings of the 2012 conference of the North American
  chapter of the Association for Computational Linguistics: Human language
  technologies}, pages 597--601. Association for Computational Linguistics.

\bibitem[Bahrainian and Dengel, 2013]{bahrainian2013sentiment}
Bahrainian, S.-A. and Dengel, A. (2013).
\newblock Sentiment analysis and summarization of twitter data.
\newblock In {\em Computational Science and Engineering (CSE), 2013 IEEE 16th
  International Conference on}, pages 227--234. IEEE.

\bibitem[Di~Iorio et~al., 2013]{di2013towards}
Di~Iorio, A., Nuzzolese, A.~G., and Peroni, S. (2013).
\newblock Towards the automatic identification of the nature of citations.
\newblock In {\em SePublica}, pages 63--74.

\bibitem[Ding et~al., 2014]{ding2014content}
Ding, Y., Zhang, G., Chambers, T., Song, M., Wang, X., and Zhai, C. (2014).
\newblock Content-based citation analysis: The next generation of citation
  analysis.
\newblock {\em Journal of the Association for Information Science and
  Technology}, 65(9):1820--1833.

\bibitem[Godbole et~al., 2007]{godbole2007large}
Godbole, N., Srinivasaiah, M., and Skiena, S. (2007).
\newblock Large-scale sentiment analysis for news and blogs.
\newblock {\em ICWSM}, 7(21):219--222.

\bibitem[Kouloumpis et~al., 2011]{kouloumpis2011twitter}
Kouloumpis, E., Wilson, T., and Moore, J.~D. (2011).
\newblock Twitter sentiment analysis: The good the bad and the omg!
\newblock {\em Icwsm}, 11:538--541.

\bibitem[Ma et~al., 2016]{ma2016improve}
Ma, Z., Nam, J., and Weihe, K. (2016).
\newblock Improve sentiment analysis of citations with author modelling.
\newblock In {\em Proceedings of NAACL-HLT}, pages 122--127.

\bibitem[Mohammad et~al., 2016]{mohammad2016semeval}
Mohammad, S., Kiritchenko, S., Sobhani, P., Zhu, X.-D., and Cherry, C. (2016).
\newblock Semeval-2016 task 6: Detecting stance in tweets.
\newblock In {\em SemEval@ NAACL-HLT}, pages 31--41.

\bibitem[Pak and Paroubek, 2010]{pak2010twitter}
Pak, A. and Paroubek, P. (2010).
\newblock Twitter as a corpus for sentiment analysis and opinion mining.
\newblock In {\em LREC}, volume~10, pages 1320--1326.

\bibitem[Pang and Lee, 2004]{pang2004sentimental}
Pang, B. and Lee, L. (2004).
\newblock A sentimental education: Sentiment analysis using subjectivity
  summarization based on minimum cuts.
\newblock In {\em Proceedings of the 42nd annual meeting on Association for
  Computational Linguistics}, page 271. Association for Computational
  Linguistics.

\bibitem[Wan and Liu, 2014]{wan2014all}
Wan, X. and Liu, F. (2014).
\newblock Are all literature citations equally important? automatic citation
  strength estimation and its applications.
\newblock {\em Journal of the Association for Information Science and
  Technology}, 65(9):1929--1938.

\bibitem[Whitelaw et~al., 2005]{whitelaw2005using}
Whitelaw, C., Garg, N., and Argamon, S. (2005).
\newblock Using appraisal groups for sentiment analysis.
\newblock In {\em Proceedings of the 14th ACM international conference on
  Information and knowledge management}, pages 625--631. ACM.

\bibitem[Wu et~al., 2015]{wu2015sentiment}
Wu, Z., Rao, Y., Li, X., Li, J., Xie, H., and Wang, F.~L. (2015).
\newblock Sentiment detection of short text via probabilistic topic modeling.
\newblock In {\em International Conference on Database Systems for Advanced
  Applications}, pages 76--85. Springer.

\bibitem[Xu et~al., 2013]{xu2013using}
Xu, H., Martin, E., and Mahidadia, A. (2013).
\newblock Using heterogeneous features for scientific citation classification.
\newblock In {\em Proceedings of the 13th conference of the Pacific Association
  for Computational Linguistics}.

\bibitem[Xu et~al., 2015]{xu2015citation}
Xu, J., Zhang, Y., Wu, Y., Wang, J., Dong, X., and Xu, H. (2015).
\newblock Citation sentiment analysis in clinical trial papers.
\newblock In {\em AMIA Annual Symposium Proceedings}, volume 2015, page 1334.
  American Medical Informatics Association.

\bibitem[Yu, 2013]{yu2013automated}
Yu, B. (2013).
\newblock Automated citation sentiment analysis: what can we learn from
  biomedical researchers.
\newblock {\em Proceedings of the American Society for Information Science and
  Technology}, 50(1):1--9.

\end{thebibliography}

\end{document}